\documentclass[conference]{IEEEtran}
\IEEEoverridecommandlockouts
\usepackage{cite}
\usepackage{amsmath,amssymb,amsfonts}
\usepackage{algorithmic}
\usepackage{graphicx}
\usepackage{textcomp}
\usepackage{xcolor}
\usepackage{multirow,multicol} 
\usepackage{bm}
\def\BibTeX{{\rm B\kern-.05em{\sc i\kern-.025em b}\kern-.08em
    T\kern-.1667em\lower.7ex\hbox{E}\kern-.125emX}}
\begin{document}

\title{Attention-based Saliency Hashing for Ophthalmic Image Retrieval}
\author{
\IEEEauthorblockN{Jiansheng Fang \IEEEauthorrefmark{1}\IEEEauthorrefmark{2}\IEEEauthorrefmark{3}, Yanwu Xu \IEEEauthorrefmark{4}, Xiaoqing Zhang \IEEEauthorrefmark{2}, Yan Hu \IEEEauthorrefmark{2}, Jiang Liu \IEEEauthorrefmark{2}\IEEEauthorrefmark{4} \thanks{Corresponding author: Jiang Liu (Email: liuj@sustech.edu.cn)}
}
\IEEEauthorblockA{
\IEEEauthorrefmark{1} School of computer science and technology, Harbin Institute of Technology, Harbin 150001, China}
\IEEEauthorblockA{
\IEEEauthorrefmark{2} Department of computer science and engineering, Southern University of Science and Technology, Shenzhen 518055, China}
\IEEEauthorblockA{
\IEEEauthorrefmark{3} CVTE Research, Guangzhou 510530, China}
\IEEEauthorblockA{
\IEEEauthorrefmark{4} The Cixi Institute of Biomedical Engineering, Chinese Academy of Sciences, Ningbo 315201, China}}

\maketitle

\begin{abstract}
Deep hashing methods have been proved to be effective for the large-scale medical image search assisting reference-based diagnosis for clinicians. However, when the salient region plays a maximal discriminative role in ophthalmic image, existing deep hashing methods do not fully exploit the learning ability of the deep network to capture the features of salient regions pointedly. The different grades or classes of ophthalmic images may be share similar overall performance but have subtle differences that can be differentiated by mining salient regions. To address this issue, we propose a novel end-to-end network, named Attention-based Saliency Hashing (ASH), for learning compact hash-code to represent ophthalmic images. ASH embeds a spatial-attention module to focus more on the representation of salient regions and highlights their essential role in differentiating ophthalmic images. Benefiting from the spatial-attention module, the information of salient regions can be mapped into the hash-code for similarity calculation. In the training stage, we input the image pairs to share the weights of the network, and a pairwise loss is designed to maximize the discriminability of the hash-code. In the retrieval stage, ASH obtains the hash-code by inputting an image with an end-to-end manner, then the hash-code is used to similarity calculation to return the most similar images. Extensive experiments on two different modalities of ophthalmic image datasets demonstrate that the proposed ASH can further improve the retrieval performance compared to the state-of-the-art deep hashing methods due to the huge contributions of the spatial-attention module.
\end{abstract}

\begin{IEEEkeywords}
Content-based Image Retrieval, Ophthalmic Image, Deep Hashing Methods, Spatial Attention, Salient Region
\end{IEEEkeywords}

\section{Introduction}
With the rapid development of radiological imaging techniques, ophthalmic images are produced in ever-increasing quantities. There are three common ophthalmic image modalities: (1) \textbf{Slit Lamp.} The silt lamp technique \cite{cheung2011validity} is usually used to examine the anterior segment and the posterior segment of the human eye, including eyelid, sclera, conjunctiva, iris, crystalline lens, and cornea. (2) \textbf{Retinal Fundus.} The retinal fundus image is a vital eye image modality that has been utilized to diagnose many ocular diseases \cite{fu2018joint,fu2018disc} by capturing the eye's inner lining and the structures of the back of the eye. (3) \textbf{ASOCT.} Anterior Segment Optical Coherence Tomography (ASOCT) is used for visualization and assessment of anterior segment ocular features, such as the tear film, cornea, conjunctiva, sclera, rectus muscles, anterior chamber angle structures, and lens \cite{hirnschall2017prediction}. In recent decades, motivated by the technique of pattern recognition and computer vision, especially deep learning methods \cite{litjens2017survey}, ophthalmic image processing plays an increasingly important role in assisting diagnosis and assessment of disease. But the objective interpretation of the ophthalmic image is fraught with high inter-observer variability and limited reproducibility. To circumvent the discrepancy between expert interpretation, prior cases with similar disease manifestations could be presented to form a reference-based assessment by Content-Based Image Retrieval (CBIR). CBIR is used for retrieving user-required images by indexing and mining visual features of image content (\textit{e.g.}, color, shape, texture) from large-scale image databases \cite{zhou2017recent}. Unlike classification and grading tasks directly giving diagnostic results, the task of ophthalmic CBIR is to improve diagnosis with evidence-based. For better assistance in assessment, CBIR should be with plenty of cases, which requires the retrieval algorithm to be both scalable and accurate. Toward this, hashing methods for CBIR arise to be a promising solution by representing a high-dimensional image as a compact hash-code due to their storage capability and processing time efficiency \cite{conjeti2017deep,wu2019deep}.

\begin{figure}[htbp]
  \centering
  \includegraphics[width=\linewidth]{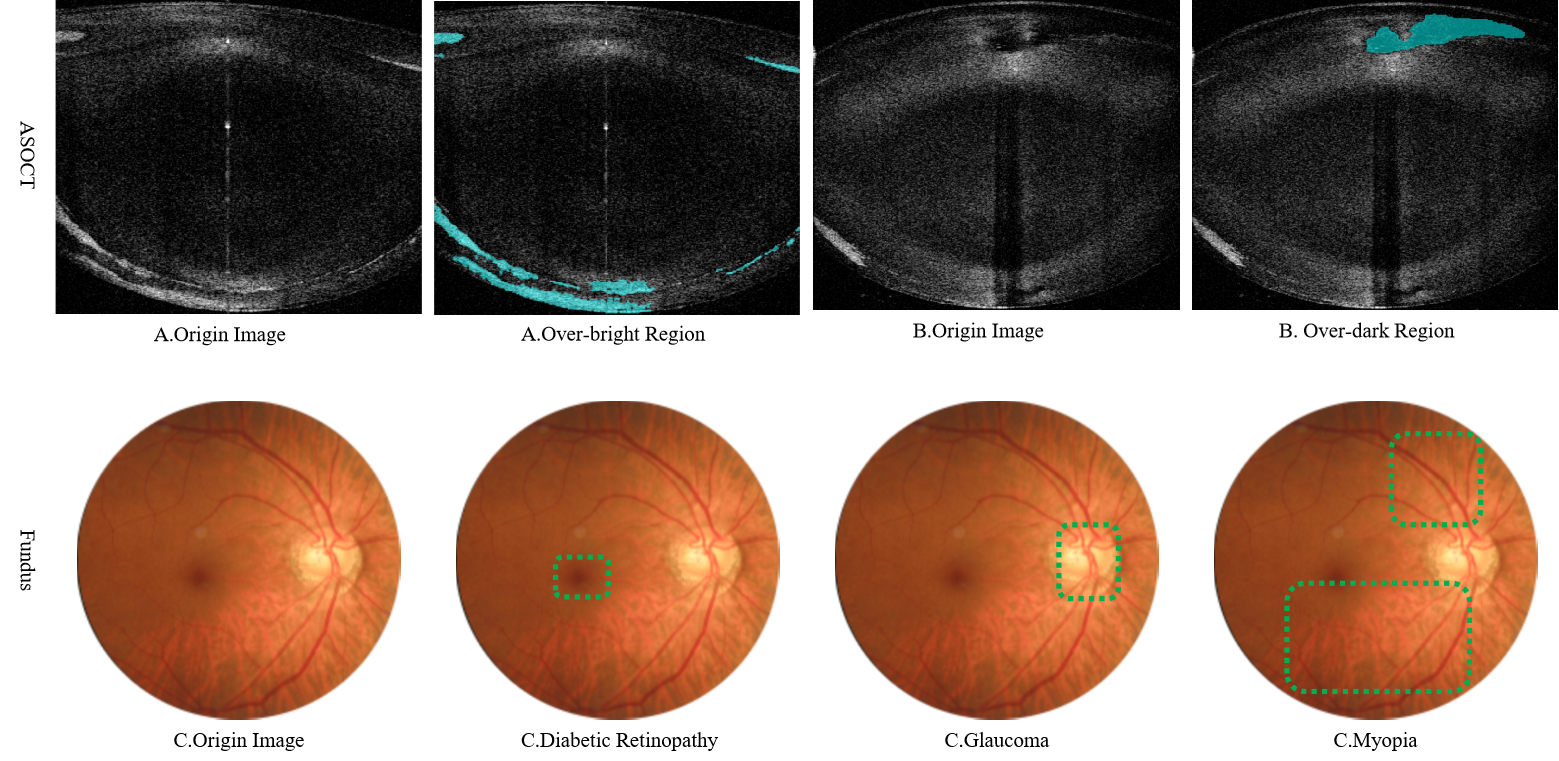}
  \caption{Schematic of salient regions in ophthalmic images.}
  \label{fig1}
\end{figure}

Recently, deep learning-based hashing methods have been widely applied for medical image retrieval, such as deep multiple instances hashing for tumor assessment \cite{conjeti2017deep}, deep residual hashing for chest X-ray images \cite{conjeti2017hashing}, order-sensitive deep hashing method for multi-morbidity medical image retrieval \cite{chen2018order}, etc. However, existing deep hashing methods do not fully exploit the learning ability of the deep network to capture the feature of salient regions pointedly. The information of salient regions is usually the main clue for clinical diagnosis. Especially in ophthalmology, the different grades or classes of ophthalmic images may be share similar overall performance but have subtle differences that can be differentiated by mining salient regions. As Fig.\ref{fig1} shows, the ophthalmic images (A) and (B) originate from ASOCT which is a non-invasive high-resolution anterior segment imaging technique. Both (A) and (B) are cataract cases with similar overall performance, but there are two types of salient regions to differentiate them: over-bright region (blue) and over-dark region (blue). Cataract grading depends heavily on detecting and observing the salient regions, so the representation of salient regions should be specially taken into consideration in ophthalmic image retrieval. On the other hand, as shown in Fig. \ref{fig1}, one retinal fundus image (C) has three symptoms that can be diagnosed according to different salient regions (green rectangle). The pathological myopic eye is usually characterized by leopard fundus, the diagnosis of Diabetic Retinopathy (DR) can rely on the spatial information of macular, and the large optic cup and abnormality of Retinal Nerve Fiber Layer (RNFL), which are typical symptoms of glaucoma. The information of these salient regions should be captured and mapped into the hash-code to play the effect of differentiating fundus images in the retrieval task. 

Inspired by the attention mechanism in the deep network \cite{woo2018cbam,chen2017sca}, to effectively mine salient regions of ophthalmic images, we propose a novel end-to-end network, named Attention-based Saliency Hashing (ASH), to promote the discriminative capability of the hash-code by exploiting salient regions. To fully mine salient regions and effectively enlarge their proportion in the hashing space, we utilize a spatial-attention module to focus on capturing the information of salient regions in the whole ophthalmic image. For similarity preservation, with the input image pairs sharing weights, we apply the siamese structure \cite{bertinetto2016fully} to learn the pairwise loss in the objective function. The pairwise distance in the hashing space is desired to be smaller if the input image pairs are similar and vice versa. The main contributions of this paper are summarized as follows:
\begin{itemize}
    \item[1)] The different grades or classes of ophthalmic images may be share similar overall performance but have subtle differences that can be differentiated by mining salient regions. According to this issue, a novel end-to-end network, named Attention-based Saliency Hashing (ASH), is proposed for ophthalmic image retrieval.
    \item[2)] With the help of a spatial-attention module, by focusing on salient regions in the whole ophthalmic image with an end-to-end manner, our ASH can boost the representation of salient regions to maximize the discriminative capability of the hash-code.
    \item[3)] The experiments on two different modalities of ophthalmic image datasets show that our ASH can significantly improve the accuracy while achieve fair efficiency, compared to the state-of-the-art deep hashing methods.
\end{itemize}
The rest of this paper is organized as follows: Section \uppercase\expandafter{\romannumeral2} introduces related works. Section \uppercase\expandafter{\romannumeral3} describes our methodology in detail. Section \uppercase\expandafter{\romannumeral4} extensively evaluates the proposed method on two ophthalmic images datasets. Section \uppercase\expandafter{\romannumeral5} gives concluding remarks.

\section{Related Works}
In this section, we introduce the most related works from two aspects: deep hashing methods and attention mechanisms.

\textbf{Deep Hashing Methods.} Existing hashing methods can be divided into data-independent methods and data-dependent methods \cite{wang2017survey,li2017losha}. The representative data-independent methods include Locality Sensitive Hashing (LSH) \cite{slaney2008locality} and its variants \cite{tang2017locality,li2017losha}. Since data-dependent methods preserve the semantic structure of the data, they usually achieve better performance \cite{jin2020deep}. The data-dependent methods, also called learning-based hashing methods, can be further categorized into \cite{chen2018order}: (1) shallow learning-based hashing methods like Iterative Quantization (ITQ) \cite{gong2012iterative}, Metric Hashing Forests (MHF) \cite{conjeti2016metric}; (2) deep learning-based hashing methods like Deep Hashing Network (DHN) \cite{zhu2016deep}, Simultaneous Feature Learning and Hashing (SFLH) \cite{lai2015simultaneous}, Deep Pairwise-Supervised Hashing (DPSH) \cite{li2015feature}, Deep Supervised Hashing (DSH) \cite{liu2016deep}. The former learns hashing functions in a two-stage manner from the hand-crafted features, which may lead to sub-optimal performance. In contrast, the latter directly tailor features for hashing through end-to-end manner with powerful Convolutional Neural Network (CNN) and has recently shown very strong performance improvements over the former \cite{zheng2017sift}. The end-to-end deep hashing methods can improve performance by using semantic information in terms of reliable class labels in a supervised way. Thus, the quality of the hash-code depends heavily on feature extraction, which is the most crucial limitation of such methods. In this work, according to the characteristics of ophthalmic images, we introduce a spatial-attention module into the network to mine salient regions.

\textbf{Attention Mechanisms.}
Recently, attention mechanisms have been successfully applied in CNNs, significantly boosting the performance of many medical image tasks \cite{oktay2018attention,nie2018asdnet}, including segmentation, recognition, and classification. For instance, an attention-based CNN \cite{li2019attention} is proposed for glaucoma detection, including an attention prediction subnet, a pathological area localization subnet, and a glaucoma classification subnet. A novel Attention Gate (AG) \cite{schlemper2019attention} can also be easily integrated into standard CNN models to leverage salient regions in medical images for various medical image analysis tasks, including fetal ultrasound classification, and 3D CT abdominal segmentation. Attention mechanisms improve the performance by guiding the model activations to be focused around salient regions. Many CNN-based saliency detection methods \cite{li2015visual,zhao2015saliency,liu2016dhsnet} consider a region and its spatial neighboring as the context to calculate the salient score of this region. Based on prior research, we argue that the attention mechanism can be beneficial to improve the performance of ophthalmic image retrieval by capturing the information of salient regions. In this work, we improve the spatial-attention module in CBAM \cite{woo2018cbam} to generate an efficient feature descriptor, without the channel-attention module.

Based on the above discussion related to the novelty of this work, the proposed ASH has two key components: (1) a ophthalmic image feature learning component with a spatial-attention module; (2) a hash-code learning component for image features with the pairwise loss. Extensive experiments on two ophthalmic image datasets demonstrate the effectiveness of our ATH.

\section{Methodology}
Our goal is to learn compact hash-codes for representing images: (a) similar images should be encoded to similar codes in the hashing space, and vice versa; (b) the information of salient regions should be prominently encoded into the discriminative image representation. Based on the spatial-attention module and siamese structure, our method trains ASH using image pairs and the corresponding similarity labels. And the pairwise loss is designed to learn similarity-preserving image representation. 

\subsection{Network Architecture}
\begin{figure*}[htbp]
  \centering
  \includegraphics[width=0.9\linewidth]{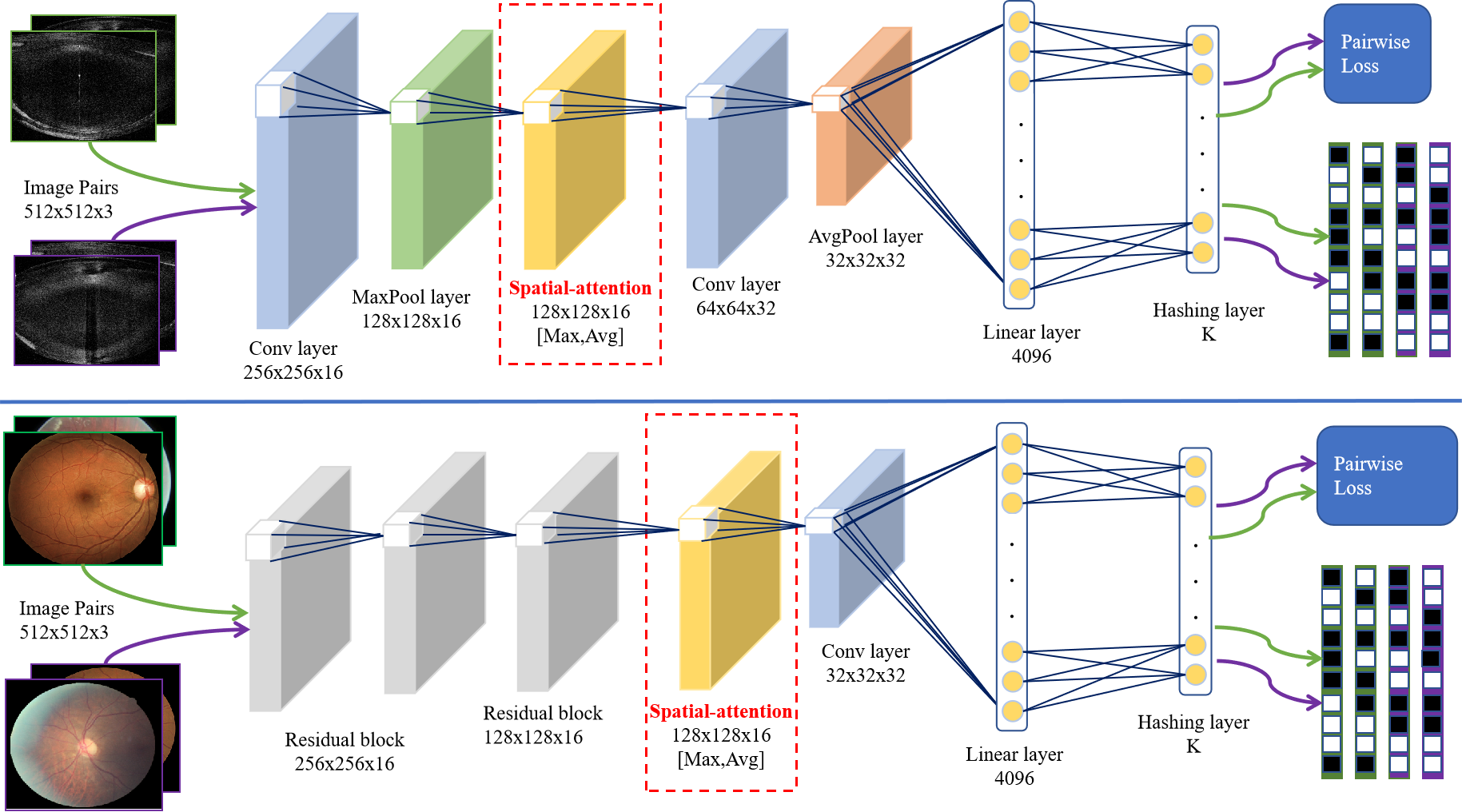}
  \caption{Illustration of the proposed ASH structure. ASH generates hash-code pairs from the input image pairs. The upper network (ASH-U) is an off-the-shelf CNN, and the lower network (ASH-L) is an application-specific CNN. A spatial-attention module is embedded into both networks.}
  \label{fig2}
\end{figure*}

As shown in Fig.\ref{fig2}, we propose a novel attention-based saliency hashing network to jointly learn visual features and the subsequent mapping to a compact hash-code. The learned visual features can effectively catch the information of salient regions. The upper network in Fig.\ref{fig2} (called ASH-U) consists of a convolutional layer and a max-pooling layer followed by a spatial-attention module and terminates in a fully-connected hashing layer for hash-code generation. And the lower network in Fig.\ref{fig2} (called ASH-L) introduces a spatial-attention module after three residual blocks \cite{he2016deep}, then a convolutional layer and terminates in a fully-connected hashing layer for hash-code generation. All the convolutional and pooling layers use \begin{math}3\times3\end{math} filters and are followed by batch normalization \cite{ioffe2015batch}. The fully-connected linear layer contains \begin{math}4096\end{math} nodes, and the fully-connected hashing layer contains \begin{math}K\end{math}-bits. All convolution layers and fully-connected layers are equipped with the ReLU \cite{nair2010rectified}. The image pairs are inputted into the ASH to generate pair hash-codes and share weights of network in training procedure. This procedure exactly utilizes the siamese structure to learn the objective function with pairwise loss for retrieval scenario.

The spatial-attention module used in this work is a improved variant of CBAM \cite{woo2018cbam} to enhance the representation of salient regions. Let \begin{math}\bm{X} \in \mathbb{R}^{C\times H\times W}\end{math} be the feature maps extracted from a convolutional layer, where \begin{math}H\end{math} and \begin{math}W\end{math} denote the height and width of each feature map, and \begin{math}C\end{math} denotes the number of feature maps (or channels) in that layer. The spatial-attention module aggregates spatial information of feature maps \begin{math}\bm{X}\end{math} by utilizing max-pooling (\begin{math}Max\end{math}) and average-pooling (\begin{math}Avg\end{math}) operations jointly along the channel axis. Both operations obtain the local maximum and average value of feature maps \begin{math}\bm{X}\end{math} respectively, then \begin{math}Max(\bm{X})\end{math} and \begin{math}Avg(\bm{X})\end{math} multiply element-wise to further weight the local salient regions. The output of the weighting operation is equipped with the sigmoid function, then multiply with the feature maps \begin{math}\bm{X}\end{math}. The spatial-attention module is defined as:
\begin{equation}
\begin{split}
\bm{X} = \bm{X} \otimes \sigma (Max(\bm{X}) \otimes Avg(\bm{X}))
\end{split}
\label{eq1},
\end{equation}
where \begin{math}Max(\bm{X})\end{math} and \begin{math}Avg(\bm{X})\end{math} are \begin{math}H\times W\times 1\end{math} dimensions, and \begin{math}\sigma\end{math} denotes the sigmoid function.

In the ASH-U, the spatial-attention module is embedded between a max-pooling layer and a convolutional layer followed by an average-pooling layer. After a max-pooling layer amplifying local salient values, the spatial-attention module can easily capture them while the value of salient regions is usually higher than the other regions. Such a spatial-attention module is beneficial to capture the information of salient regions in the high-resolution ophthalmic image. In the view of flexibility and usability, in the ASH-L, we further incorporate the spatial-attention module into the application-specific CNN structures due to their different capability of feature extraction. The learning procedure of our ASH focuses on the feature extraction of salient regions on the raw pixels of input images by using an attention-based CNN. Such hierarchical non-linear function exhibits a powerful learning capacity and encourages the learned feature to capture the information of salient regions by introducing the attention mechanism. 

\subsection{Pairwise Loss}
The main idea behind the pairwise loss of siamese structure is that the codes of a pair of samples from the same class should be as close as possible, while the codes of a pair of samples from the different classes being far away. Based on this objective, the loss function is naturally designed to minimize the distance of similar image pairs and maximize the distance of dissimilar image pairs. 

Mathematically, giving a set of samples with \begin{math}2N\end{math} length and corresponding class labels \begin{math}\bm{L}=\{1,\dots,C\}\end{math} and dividing the samples with labels into two equal parts \begin{math}\bm{I_{1}}=\{\bm{x_{1,M}},\dots,\bm{x_{N,M}}\}\end{math} and \begin{math}\bm{I_{2}}=\{\bm{x_{1,M}},\dots,\bm{x_{N,M}}\}\end{math}, we typically define similar labels \begin{math}\bm{Y}=\{y_{1},\dots,y_{N}\}\end{math}, where \begin{math}y_{i}=0\end{math} if \begin{math}x_{i,M}\end{math} in \begin{math}\bm{I_{1}}\end{math} and \begin{math}\bm{I_{2}}\end{math} are the same class label (similar) and \begin{math}y_{i}=1\end{math} otherwise (dissimilar). ASH aims at learning a mapping from input image pairs to hash-code pairs \begin{math}\bm{H_{1}}=\{\bm{h_{1,K}},\dots,\bm{h_{N,K}}\}\end{math} and \begin{math}\bm{H_{2}}=\{\bm{h_{1,K}},\dots,\bm{h_{N,K}}\}\end{math}. For scalable retrieval, the hash-code bits \begin{math}K\end{math} is much smaller than the dimension of ophthalmic image \begin{math}M\end{math}. The pairwise loss with respect to image pairs is defined as:
\begin{equation}
\begin{split}
L(\bm{H_{1}},\bm{H_{2}},\bm{Y}) & =  \\
  & \frac{1}{2}(\bm{1}-\bm{Y})D(\bm{H_{1}},\bm{H_{2}}) + \\
  & \frac{1}{2}\bm{Y}\max\{r\cdot K - D(\bm{H_{1}},\bm{H_{2}}),0\}
\end{split}
\label{eq2},
\end{equation}
where \begin{math}D(\cdot,\cdot)\end{math} denotes \begin{math}L_{2}\end{math} normalization to measure the distance between hash-codes, \begin{math}\textbf{1}\end{math} is a vector of all ones, and \begin{math}r\in[0,1]\end{math} is a weighting parameter that controls the punish strength of differentiating degrees between dissimilar images. In Eq. (\ref{eq2}), the first term punishes the similar images mapped to far hash-code, and the second term punishes the dissimilar images mapped to close hash-code when their distance falls below the margin threshold \begin{math}r\cdot K\end{math}.

In the loss function of ASH, the contrastive loss form is applied as that only those dissimilar pairs having their distance within a radius are eligible to contribute to the loss function. When \begin{math}r=0\end{math}, there is no punishment of dissimilar images mapped to close hash-code. When \begin{math}r=1\end{math}, the hash-code of dissimilar images is required to be complete difference. If \begin{math}r=0.5\end{math}, this implies that half of the hash-code bits between dissimilar images is required to be different. To be especial, we do not impose an additional regularizer on the loss function so that we can independently observe the contributions of the spatial-attention module.

\section{Experiments}
We perform extensive experiments on two ophthalmic image datasets to prove: (1) the salient regions of ophthalmic images can improve discriminative ability of the hash-code, and (2) the spatial-attention module can effectively mine the salient regions of ophthalmic images.  

\subsection{Ophthalmic Datasets}
(1) \textbf{ASOCT-Cataract.} In the ASOCT-Cataract dataset, each image is labeled as coprresponding grade according to the degree turbidity of cortex, nucleus, subcapsular. Among the three types, the size of the subcapsular is smaller than the others. For validating the capability of our ASH in capturing the information of salient regions, Oculus Dextrus (OD) images labeled five-level turbidity of subcapsular are chosen as training samples (7,000) and query images (700). (2) \textbf{Fundus-iSee.} The Fundus-iSee dataset with four disease classification consists of 10,000 high-resolution images labeled by professional doctors with rich clinical experience, having 720 images of Age-related Macular Degeneration (AMD), 270 images of DR, 450 images of glaucoma, 790 images of myopia, 7770 images of normal. We randomly extract ten percentage of each type for query test, 1,000 images. As shown in Fig. \ref{fig1}, the performance of ASOCT images grading and fundus images classification depend heavily on the differentiation of salient regions. The selected two ophthalmic datasets can be used to verify the effectiveness of our ASH in mapping the information of salient regions into the hash-codes.

\subsection{Evaluation Settings}
\begin{table}[!t]
\renewcommand{\arraystretch}{1.0}
\caption{mAP of ASH over the varying weighting parameter $r$, hash-code bits $K$ and top-10 returned list on the ASOCT-Cataract and Fundus-iSee datasets.}
\begin{center}
\begin{tabular}{ |c|c|c|c|c|c| } 
    \hline 
    \textbf{Datasets} & \textbf{$r$}  & $K=12$ & $K=24$ & $K=36$ & $K=48$\\
        \hline 
        \multirow{3}*{ASOCT-Cataract}
        & $0.3$ & 0.5918 & 0.7001 & 0.5261 & 0.5404  \\
        & $0.5$ & 0.7000 & 0.8133 & 0.6230 & 0.8161  \\
        & $0.7$ & 0.5047 & 0.6444 & 0.7439 & 0.8856  \\
        \hline 
        \multirow{3}*{Fundus-iSee}
        & $0.3$ & 0.6470 & 0.5057 & 0.5462 & 0.6226  \\
        & $0.5$ & 0.6346 & 0.5678 & 0.5273 & 0.7233  \\
        & $0.7$ & 0.6243 & 0.6205 & 0.6771 & 0.7528  \\
        \hline 
    \end{tabular}
 \end{center}\label{tb1}
\end{table}
In our comparative study, we used a data-independent method: \textbf{LSH} \cite{slaney2008locality}; a shallow learning-based hashing method: \textbf{MHF} \cite{conjeti2016metric}; three deep learning-based hashing methods: \textbf{DPSH} \cite{li2015feature}, \textbf{DSH} \cite{liu2016deep}, \textbf{DRH} \cite{conjeti2017hashing}. In our implementation of Fig.\ref{fig2}, the experiment on ASOCT-Cataract used ASH-U, and the experiment on Fundus-iSee used ASH-L. Further, to study the effects of different CNN structures, we embedded the spatial-attention module before the fully-connected layer of DPSH and DSH, called \textbf{DPSH-A} and \textbf{DSH-A} respectively, and our ASH-L is the variant of DRH by introducing the spatial-attention module. To embody the advantage of the spatial-attention module, we used the same pairwise loss for all deep hashing methods. According to the pairwise loss, the weighting parameter \begin{math}r\end{math} and the hash-code bits \begin{math}K\end{math} are synchronous to each other. Reasonably, setting \begin{math}r=0.5\end{math} refers to that half of the hash-code bits between dissimilar images should be different. As shown in Table \ref{tb1}, with the hash-code lengthen, \begin{math}r\end{math} can be correspondingly set higher than the short hash-code, then the performance can correspondingly improve at the cost of storage and search efficiency. As a trade-off between performance and search cost, we set \begin{math}r=0.5\end{math} in our experiments, then we evaluated the performance over hash-code bits from 12, 24, 36 to 48, and the most similar images from 5, 10, 15 to 20.

Three metrics were adopted to measure the precision and retrieval quality in our experiments. (1) \textbf{Mean Hit Ratio (mHR).} HR is designed to measure how many images in the returned list are similar to the query image. (2) \textbf{Mean Average Precision (mAP).} In the returned list, AP averages the rank positions of images similar to the query image to measure the rank quality. (3) \textbf{Mean Reciprocal Rank (mRR).} RR refers to the reciprocal of the ranking of the first similar image in the returned list. Our ASH is implemented under PyTorch framework and experiments are run on Geforce RTX 2080 Ti. The indexing and similarity calculation for evaluation uses Faiss \cite{johnson2019billion} which is a library for efficient similarity search and clustering of dense vectors. We use the mini-batch stochastic gradient descent with 0.9 momentum. The mini-batch size of images is fixed as 10 and the weight decay parameter as 0.001. The input image pairs were randomly sampled in every iteration. All deep models are trained from scratch, setting the iteration number as 50. The parameters of comparative methods are set according to their implementation details in the corresponding works, and the best performance is reported. 

\subsection{Results and Analysis}
\begin{figure*}[htbp]
  \centering
  \includegraphics[width=\linewidth]{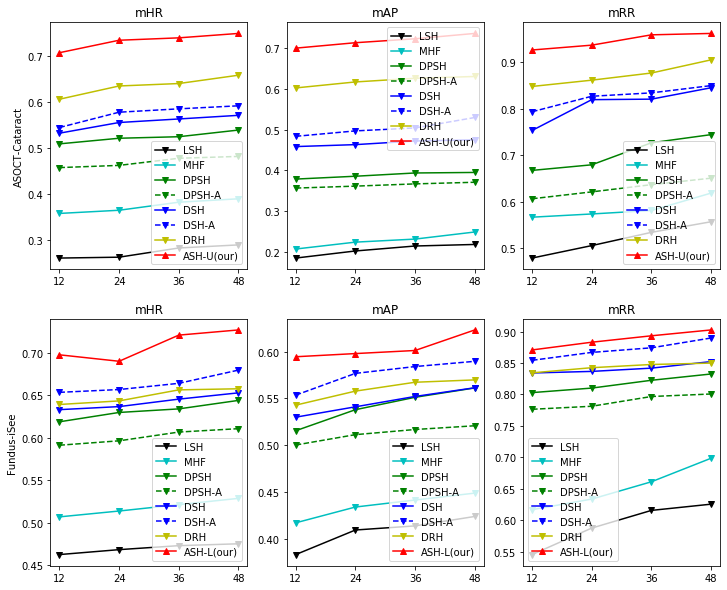}
  \caption{mHR, mAP, mRR of the ASOCT-Cataract and Fundus-iSee datasets w.r.t varying the hash-code bits over top-10 returned list.}
  \label{fig3}
\end{figure*}
Fig.\ref{fig3} shows the retrieval performance of different hashing methods in terms of mHR, mAP, and mRR for different hash-code bits. Due to the contributions of the spatial-attention module, the performance of our ASH (red line) is more superior to other state-of-the-art hashing methods: (a) data-independent method (LSH-black line) and shallow learning-based hashing method (MHF-cyan line) using hand-crafted features, (b) deep learning-based hashing methods without attention mechanisms (DPSH-solid green line, DSH-solid blue line, DRH-solid yellow line). Although the longer bits of hash-code can improve the performance due to containing more information of images, we have to make a balance between accuracy and efficiency. The hash-code bits can be properly enlarged according to the volume of the dataset increasing. On the one hand, compared to the deep learning-based hashing methods using our spatial-attention module but different network structures, our ASH consistently outperforms both DPSH-A (dotted green line) and DSH-A (dotted blue line) by 13\%-31\% on mHR, 11\%-21\% on mAP, 9\%-17\% on mRR. On the other hand, we can observe that the performance of DPSH (solid green line) outperforms DPSH-A (dotted green line) while the latter embeds our spatial-attention module. This demonstrate that the effectiveness of attention mechanisms relates to the network structure. For the classification dataset (Fundus-iSee), our ASH-L, which is a variant of DRH adding spatial-attention, is applied to extract visual features. And the order of performance is ASH-L$>$DSH-A$>$DRH$>$DSH. But in the grading dataset (ASOCT-Cataract), the order of performance is ASH-U$>$DRH$>$DSH-A$>$DSH while our ASH-U adopts an off-the-shelf CNN structure. Compared to the different diseases (classification), the region feature is more difficult to extract in the different degrees (grading) of the same disease. The above analysis demonstrates that the structure of our ASH-U can further improve the performance for the retrieval task of grading datasets compared to DRH, while the structure of our ASH-L and DRH is more suitable for the retrieval task of classification datasets than DSH. Through experiments on the ASOCT-Cataract and Fundus-iSee datasets, our ASH is proved to be the best performance by embedding the spatial-attention module into the specific network structure. 

\begin{figure*}[htbp]
  \centering
  \includegraphics[width=\linewidth]{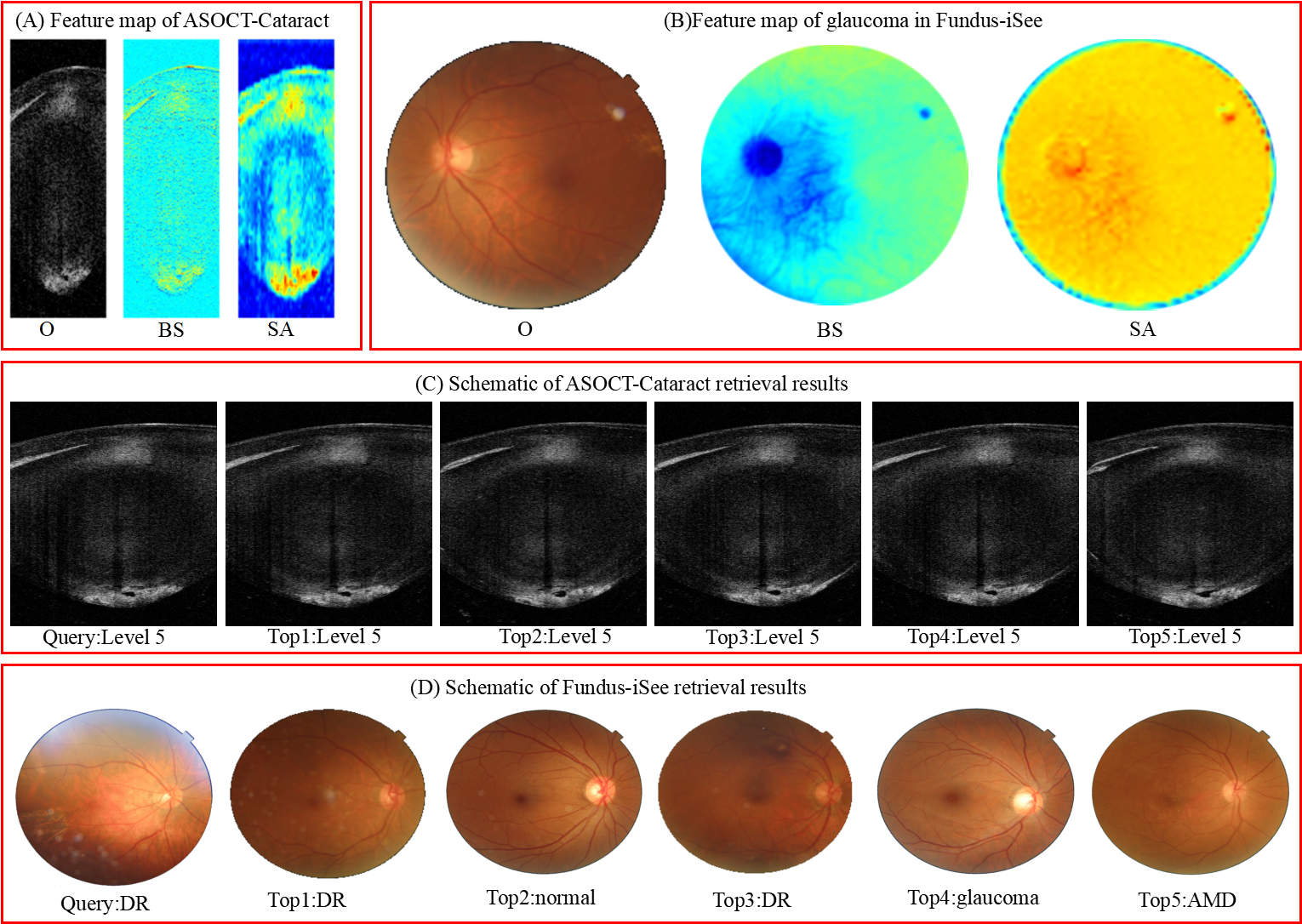}
  \caption{Qualitative results of our ASH: (A) Feature map of ASOCT-Cataract; (B) Feature map of glaucoma in Fundus-iSee;  (C) Schematic of ASOCT-Cataract retrieval results; (D) Schematic of Fundus-iSee retrieval results. }
  \label{fig4}
\end{figure*}
In Fig.\ref{fig4}, to qualitatively show the contribution of the spatial-attention module, the feature maps of ASOCT-Cataract and Fundus-iSee are shown in (A) and (B). In both (A) and (B), O is the original image, BS and SA are the feature map of the convolutional layer before and after the spatial-attention module, respectively. We can observe: (1) the over-bright salient regions of subcapsular are captured in image SA of (A); (2) large optic cup can be suspected from image SA of (B). This demonstrate that our ASH can detect the subtle difference of ophthalmic images by mining the salient regions with the help of the spatial-attention module. The visual information of salient regions are learned in our ASH and are mapped into the hash-code for differentiating ophthalmic images. The higher mAP means more similar images in the returned list are ranked ahead and can help users to find the required disease case quickly. As (C) of Fig.\ref{fig4} shows, for the ASOCT-Cataract dataset, the top-5 most similar images and the query image are all turbidity level 5 of subcapsular. In (D) of Fig.\ref{fig4}, for the Fundus-iSee dataset, disturbed by the similar overall performance of different classes, the top-5 most similar images are different classes by querying DR, but the first and third images are DR. Through the classification and grading on two different ophthalmic image modalities, our ASH is proved to be powerfully capable of capturing the information of salient regions and differentiating them. In the feed-warding process, the spatial-attention module catches the spatial information of salient regions. And in the feed-backing process, the class or grade semantic information are learned by the pairwise loss and encoded into the salient regions. Thus, salient regions not only be captured by the spatial-attention module, but also be differentiated by the pairwise loss according to their corresponding class or grade.

\begin{table}[!t]
\renewcommand{\arraystretch}{1.0}
\caption{mAP of the ASOCT-Cataract and Fundus-iSee datasets w.r.t varying number of the returned list over hash-code bits of 12.}
\begin{center}
\begin{tabular}{ |c|c|c|c|c|c| } 
    \hline 
    \textbf{Datasets} & \textbf{Methods}  & \textbf{top-5} & \textbf{top-10} & \textbf{top-15} & \textbf{top-20}\\
        \hline 
        \multirow{4}*{ASOCT-Cataract}
        & DPSH & 0.4368 & 0.3785 & 0.3462 & 0.3262  \\
        & DSH & 0.4950 & 0.3583 & 0.2888 & 0.2476  \\
        & DRH & 0.5920 & 0.5023 & 0.4477 & 0.4099 \\
        & ASH & \textbf{0.7985} & \textbf{0.7000} & \textbf{0.6240}  & \textbf{0.5706} \\
        \hline 
        \multirow{4}*{Fundus-iSee}
        & DPSH & 0.5362 & 0.5154 & 0.5034 & 0.4983  \\
        & DSH & 0.5664 & 0.5399 & 0.5301 & 0.5253  \\
        & DRH & 0.5716 & 0.5428 & 0.5326 & 0.5288 \\
        & ASH & \textbf{0.6490} & \textbf{0.6346} & \textbf{0.6272} & \textbf{0.6233} \\
        \hline 
    \end{tabular}
 \end{center}\label{tb2}
\end{table}
Based on the above analysis, the effectiveness of our ASH is verified by observing the quantitative result in Fig. \ref{fig3}, and the contributions of the spatial-attention module is confirmed by observing the qualitative result in Fig. \ref{fig4}. Next, we introduce the performance over different numbers of the returned list different and analyze the efficiency of our ASH. As shown in Table. \ref{tb2}, compared to the three deep hashing methods, our ASH all achieve the best performance over varying number of returned lists from 5, 10, 15 to 20. When the returned list lengthens, the performance of all methods declines to some extent. The better performance of longer returned list implies more powerful capability of differentiating similar images based on the similarity calculation of hash-codes. This further confirm the contributions of the spatial-attention module. At the last analysis of experiments, the efficiency of the proposed method will be discussed on four-folds by putting the Fundus-iSee dataset as an example. (1) \textbf{Feature computation time.} Based on the pre-trained ASH model with 12-bits hash-code, the feature extraction of the training set of 9000 images can be completed in 33 seconds by using GPU. (2)\textbf{Retrieval time.} After features mapping into 12-bits hash-code, the index of the training set is built in 1 second by using Faiss. Then the retrieval of the test set of 1000 images can be done in 3000 $ms$ by returning top-10 most similar images. (3) \textbf{Training time.} The training process takes about one hour by setting the iteration number as 50. (4)\textbf{Memory cost.} During model training with a batch size of 10, the memory-consuming is about 2000 Mbps. The online search for the index also consumes about 3000 Mbps. Compared to three deep hashing methods, the complexity of our ASH is slightly higher in training time and memory cost due to the added spatial-attention module and is fair in feature computation time and retrieval time. According to the above analysis of efficiency, our ASH can provide fair retrieval responses with significantly improving the performance, compared to the state-of-the-art deep hashing methods.

\section{Conclusions}
To address the differentiating difficulty of ophthalmic images, the Attention-based Saliency Hashing (ASH) embed a spatial-attention module into the network structure to focusing on the feature extraction of salient regions. By aggregating and weighting the spatial information of salient regions, our spatial-attention module can boost the effect of highly active spatial response in ophthalmic images with the help of the pairwise loss. The experiments on two different ophthalmic image modalities, including fundus and ASOCT, demonstrate that our ASH could obtain state-of-the-art performances in ophthalmic image retrieval. Further analysis confirmed that the representations of salient regions make significant contributions to boost the performance in the experiments. The results also show that our ASH can meet the accurate and scalable requirements to assist the evidence-based diagnosis. In future work, we will keep focusing on the feature extraction of salient regions. The representation of salient regions is a promising task in the community of ophthalmic image retrieval. 

\bibliographystyle{IEEEtran}
\bibliography{main}

\end{document}